\documentclass[10pt, apacite]{article}
\usepackage[utf8]{inputenc}
\usepackage[T1]{fontenc}
\usepackage{graphicx}
\usepackage{caption}
\usepackage{subcaption}
\usepackage{amsmath}
\usepackage{setspace}
\usepackage[margin=1.0in]{geometry}
\usepackage{natbib}
\usepackage{longtable}
\usepackage{url}

\title{A Robotic Prosthesis for an Amputee Drummer}

\author{Mason Bretan\\
\texttt{Georgia Tech}\\
\texttt{Atlanta, GA}\\
\and
Deepak Gopinath \\
\texttt{Georgia Tech}\\
\texttt{Atlanta, GA}\\
\and
Philip Mullins \\
\texttt{Meka Robotics}\\
\texttt{San Francisco, CA}\\
\and
Gil Weinberg \\
\texttt{Georgia Tech}\\
\texttt{Atlanta, GA}\\
}
\date{December 2016}

\begin{document}
\maketitle

\abstract{
The design and evaluation of a robotic prosthesis for a drummer with a transradial amputation is presented. The principal objective of the prosthesis is to simulate the role fingers play in drumming. This primarily includes controlling the manner in which the drum stick rebounds after initial impact. This is achieved using a DC motor driven by a variable impedance control framework in a shared control system. The user's ability to perform with and control the prosthesis is evaluated using a musical synchronization study. A secondary objective of the prosthesis is to explore the implications of musical expression and human-robotic interaction when a second, completely autonomous, stick is added to the prosthesis. This {\it wearable robotic musician} interacts with the user by listening to the music and responding with different rhythms and behaviors. We recount some empirical findings based on the user's experience of performing under such a paradigm.
}

%*****************************************************************************************************
\section{Introduction}
As robots become more pervasive in everyday life and begin to function outside of purely industrial environments, it is essential for methods that support fluid and intuitive human-robotic interactions to be designed and implemented. Such methods must enable a natural flow of information between human and robot in both the cognitive and physical domains.

{\it Wearable robotics}, such as prosthetics, orthotics, and exoskeletons, provide a unique and interesting platform for research in robotic interaction and control because of the intimate user interface. These systems may extend, enhance, or replace limbs \citep{herr2003cyborg} and in some cases may even equip the user with an additional limb (such as a third arm or additional fingers) \citep{llorens2012demonstration}. Typically, the wearable robot attempts to understand a person's intent through some means of sensing including button presses, electromyography (EMG), and electroencephalography (EEG) allowing it to behave according to the user's desires. These robotic behaviors and functions assist the user by providing capabilities that may have been lost due to amputation or were never present as a result of natural human anatomy.

In this article we describe a wearable robotic prosthetic designed for an amputee drummer. Electromechanical prosthetics are comprised of sensing and control technology and designed to recreate natural human function. Dellon and Matsuoka describe three areas of challenge when attempting to replicate human function using prosthetics: electromechanical implementation, extraction of intent, and interface design or usability \citep{dellon2007prosthetics}. The demands and difficulties inherent to these areas are significantly amplified when designing a device that enables performance on a musical instrument. The relationship between a musician and his or her musical instrument is one of extreme intimacy and requires a level of chemistry that is not commonly apparent or necessary in other human-interface interactions.

Musical virtuosity is dependent on increased dexterity, accurate timing, and subtle control. Permitting an amputee the ability to attain such virtuosic control utilizing the current state-of-the-art sensing and mechanical technology is extremely challenging. As a result, our first attempt at developing a robotic drumming arm falls short of a perfect emulation of a human drumming arm, however, we see many successes in our methods and great promise to expand upon and improve the technology. Additionally, our wearable robotic system not only recreates, but also enhances natural human ability through aspects including increased speed and endurance. The potential for new and interesting music that leverages such augmented abilities can lead to novel ways for musicians (disabled and able-bodied alike) to incorporate robotics in performance.

In the first part of this paper we describe our motivation, design, and evaluation of the robotic drumming arm. The primary objective is to recreate the functionality lost due to amputation. The system employs a variable impedance control framework that requires the human and robot to work together to achieve some desired musical output. We evaluate the system through experiments requiring performance and completion of musical synchronization tasks.

Much of our research focuses on robotic musicianship -- the design and construction of robots capable of understanding and performing music in interactive scenarios such that the human and machines share control over the final musical output. In the second part of this paper we describe the incorporation of an additional stick to the prosthetic. The second drumstick functions as a robotic musician and responds to the music and improvises based on computational music analysis.

While the amputee drummer does not have full control over the second stick, he can react and collaborate with it, creating novel interactive human-robot collaboration. In designing the algorithms controlling both drum sticks, we have taken into consideration the shared control aspects that the human and the machine have over the final musical result. We also consider the physical inter-dependencies that may occur due to the fact the robotic musician is an extension of the human's own body. While the artificial intelligence is designed to generate different stroke types and rhythmic patterns, the human drummer can influence the final musical result through his own motion and manipulation of the distance between the autonomous stick and the drum. Concurrently, the autonomous stick has the opportunity to complement, contrast, or disrupt the drummer's own patterns.

Such a wearable system yields several research questions: What type of interaction between the human and robot is optimal for achieving fluid and comfortable playability that encourages musical cooperation, yet remains non-restrictive? What type of sensing by the robot is necessary to support creative musicianship? Is it possible for other musicians in the group to interact with and influence the robotic musician?

\section{Related Work}

\subsection{Shared Control in Robotics}
According to Schirner et al. most robots can be categorized into one of two categories determined by the level of human-robot interaction involved: 1) Robots that are fully autonomous performing specific tasks, such as industrial robots working on factory assembly lines and 2) robots that are tele-operated with minimal artificial intelligence capabilities that are considered to be passive tools entirely controlled by humans \citep{schirner2013future}. These categories describe the two extremes of which humans and robots interact, however, in many cases a more interdependent relationship between human and machine is desired and a shared control paradigm is adopted. Under a shared control framework a synergistic collaboration between the human and the robot exists in order to achieve a goal or complete a task in such a way that the cognitive and physical load on the human is significantly reduced.

Shared control robotic systems are adopted in several domains including assistive technology, teaching environments, and even more artistic and creative enterprises. Some human-wheelchair systems employ such a shared control framework to safely navigate an obstacle filled environment while accounting for variability in user performance \citep{li2011dynamic}. Tele-operated surgical robots can be used as training tools for novice surgeons through a paradigm in which the control is shared between mentor and trainee. The haptic feedback forces provided to each surgeon is inversely proportional to the control authority. The relative expertise levels also determine the degree to which the motion of the robot is affected when the individual commands are issued, thereby, reducing the chances of an inadvertent error \citep{nudehi2005shared}. Robotic musicianship research focuses on exploring new avenues of musical expression through human-robotic interaction \citep{weinberg2006toward}. The artistic premise being robots that perform and generate original music can push humans to unknown musical white spaces inspiring creativity and even the development of new genres.

Within the shared control paradigm itself two categories \citep{abbink2010neuromuscular} are usually identified based on the type of human robot interaction involved: 1) {\it Input mixing shared control} in which the user provides input to the robot and the robot provides input to the underlying control system and 2) {\it haptic shared control} in which the user and robot directly provide input to the control system. The system we present in this work incorporates elements of both paradigms, allowing the user to influence the behavior of the robot as well as directly influencing the final musical product.

A common feature found among different shared control systems is the presence of a closed feedback control loop. The feedback can take a number of forms including visual \citep{gentili2013human}, haptic \citep{nudehi2005shared}, and auditory \citep{kapur2011machine,maes2011man}. Often, a system incorporates multiple modes of feedback \citep{cakmak2010designing,hoffman2010shimon}. State-of-the-art prostheses generally provide visual and haptic feedback to the user \citep{cipriani2008shared}. However, despite the advantages of shared control in these applications, it is important for the users to have the capability to initiate a passive mode during situations when the robot behavior becomes intrusive or if the goal state is not reached successfully or changes midway \citep{gentili2013human}.

\subsection{Wearable Robotic Limbs}

Advances in areas of material sciences, microprocessor technology, and pattern recognition have encouraged the development of reliable and easy to use upper extremity prosthetic technology \citep{lake2006progressive}. In such technology a shared control framework is used to reduce the amount of attention required of the user \citep{cipriani2008shared}. Much of the current research in upper-limb prosthetics focuses on improving the performance of grasping \citep{cipriani2009progress}. Though several types of control mechanisms exist for assistive technologies such as brain-machine interfaces \citep{kim2006continuous,hochberg2006neuronal} and body powered prostheses, almost all of the state-of-the-art commercially available hand prostheses use EMG signals from the residual amputation stump for control of electromechanical joint movement (ProDigits and iLimb from Touch Bionics, Otto Bock myoelectric prosthesis). EMG signals are considered to be the most physiologically natural and the greatest advantage of the myoelectric prosthesis is the increased grasp strength that it provides \citep{herr2003cyborg}.

Wearable robotics are not only used for amputees or the physically impaired, but are also used for augmenting the abilities of natural human anatomy. Supernumerary limbs and fingers help to reduce load on the body and can provide additional support in completing specific tasks \citep{wu2014bio, davenport2013supernumerary}. Unlike most commonly seen wearable prostheses, which exhibit an input mixing shared control paradigm, these supernumerary robotic limbs (SRLs) exhibit a haptic shared control paradigm \citep{llorens2012demonstration}. The person and robot coordinate their efforts while both directly influencing the interface.

The objectives of our work require functionalities that support both shared control paradigms. In the remaining sections we describe these objectives and our methods for attaining them.

\section{A Robotic Drumming Prosthesis}
Prostheses come in many different forms depending on the users' needs. When developing a prosthetic, designers consider the location of the amputation, the necessary functions the device should be able to perform, the ideal communication interface (such as EMG, EEG, potentiometers, buttons, etc.), and any physical constraints that if failed to be addressed may render the device to be more hindering rather than assisting (such as weight, size, shape, and latency). In this section we discuss our drumming prosthesis in detail.

\subsection{Motivation}
Jason Barnes is a drummer whom, due to an unfortunate accident, required a transradial amputation (below the elbow through the forearm). This means the degree-of-freedom (DoF) provided by the elbow is present, but the DoFs offered by the wrist and fingers have been lost. The nature of such an amputation requires a novel prosthetic design that provides a functionality that is not typically addressed by designers of stand-alone robotic percussionists.

The methods and strategies of current robotic drummers are not applicable for Jason because most robotic drumming machines use solenoid based systems that provide the necessary vertical motion for striking a drum \citep{singer2004lemur, kapur2011machine, maes2011man, weinberg2006robot}. In these systems a single DoF is sufficient for a wide range of musicality and the speeds at which the solenoids are capable of moving greatly exceed that of a human drummer. This method is equivalent to a human using only his elbow to generate each drum strike. In reality, human drummers use a combination of several DoFs provided by the elbow, wrist, and fingers allowing for many types of strokes producing different sounds and speeds (see figure 1). 

\begin{figure}[h!]
\center
\includegraphics[width=.5\textwidth]{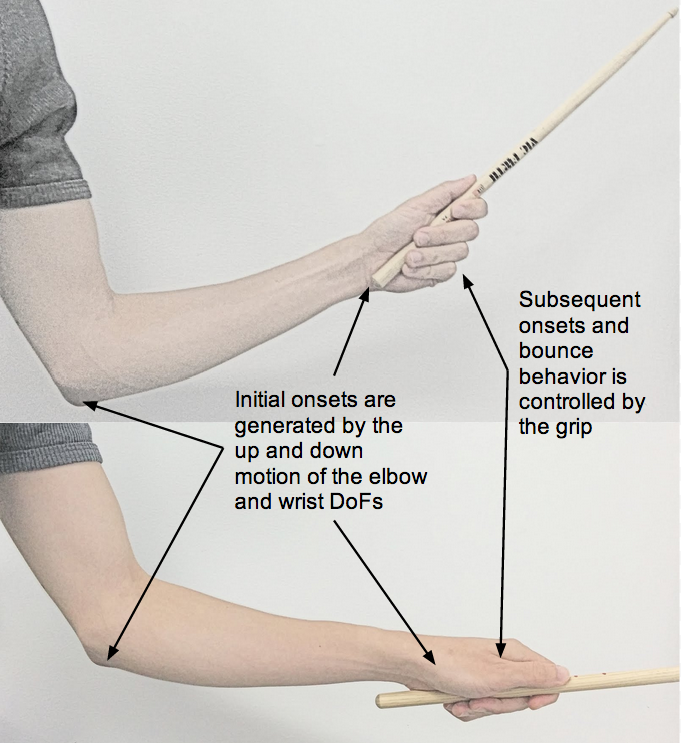}
\caption{}
\label{arm_dofs}
\end{figure}

For two hits separated by a large temporal interval a drummer uses a single-stroke in which the elbow moves up and down for each strike (similar to a single DoF system employed by robots). For two hits separated by a very small temporal interval a drummer may use a {\it double-stroke} in which the elbow moves up and down once for the first strike and the second strike is achieved by adjusting the fingers' grip on the drumstick to generate a bounce. Single-strokes generated by solenoids and motors can move at such great speeds that it is not necessary to create double-stroke functionality. Though a double-stroke is considered a fundamental rudiment of drumming it is a challenging task to recreate in robotics. Using fast single-strokes with solenoids simplifies the engineering, while achieving similar results to double-strokes.

It is possible to use the more common single solenoid system allowing us to disregard bounce, however, in order to leverage Jason's previous knowledge of drumming and take advantage of his residual elbow DoF it is important to model natural human function as close as possible. The system should be designed in such a way that it allows Jason to make use of his elbow in a manner in which he is familiar and that is typical of able-bodied drummers. Therefore, the primary objective is to develop a prosthetic that enables the initial onset of a strike to be generated using the elbow DoF and then simulates the role of the fingers to control the rebound for subsequent strikes.

\subsection{Design}
\subsubsection{Primary Goals}
There were numerous considerations that needed to be addressed in designing the drumming prosthesis. Unlike many myoelectric prostheses we chose not to use an anthropomorphic design. It is imperative that the device be robust enough to handle frequent large impact loads inherent to drumming. Furthermore, in order to reduce unnecessary muscle fatigue the prosthesis needs to be as light as possible with a center of gravity close to the elbow to reduce rotational inertia. An original non-anthropomorphic design allowed us to reduce the amount of hardware necessary and control for the center of mass. It was crucial to find a balance between lightweight design and high performance without creating something overly complex so that the cognitive and physical loads of operating the device were manageable. Therefore, we attempted to create finger and grip functionality using a single motor that would drive the tension (or rebound behavior) of a mounted drum stick (see figure 2).

\begin{figure}[ht]
\center
\includegraphics[width=.8\textwidth]{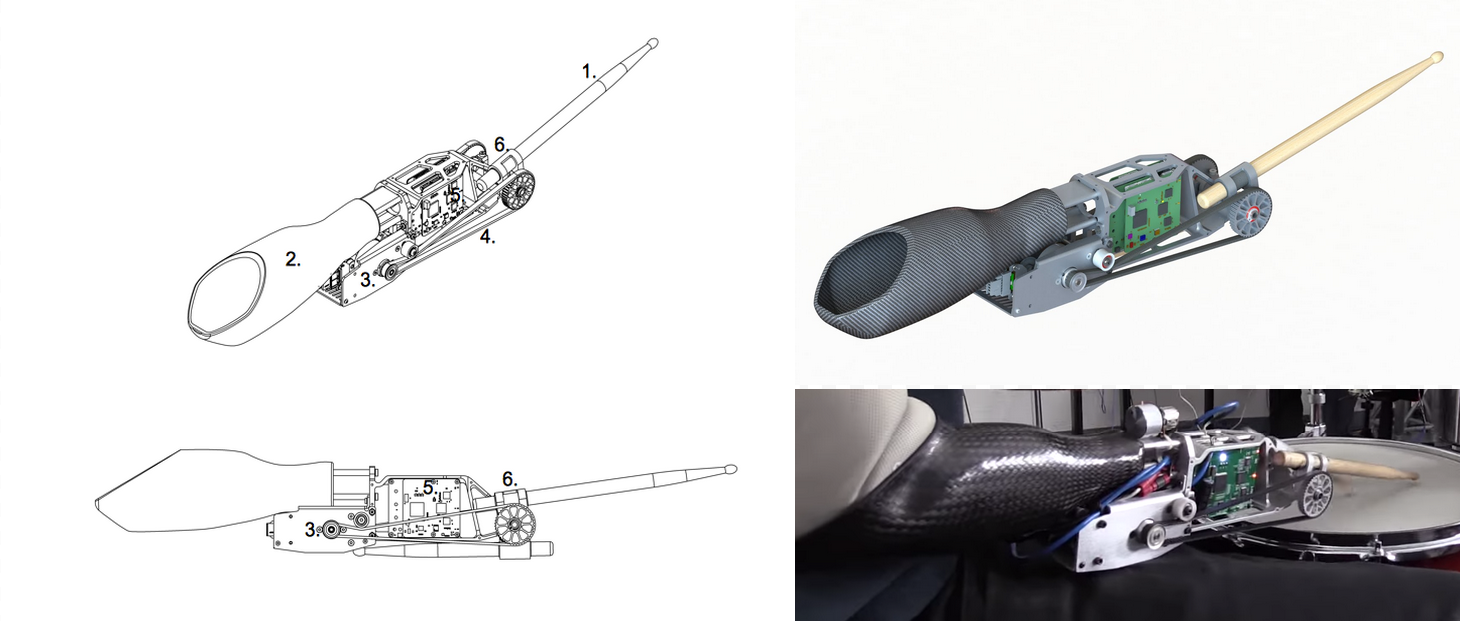}
\caption{{\bf 1.} drum stick {\bf 2.} socket {\bf 3.} casing for motors {\bf 4.} belt connecting motor to drum stick mount {\bf 5.} processor {\bf 6.} drum stick mount}
\label{primary}
\end{figure}

\subsubsection{Motors}
A large multi-stage gearbox, typical of robotic actuators, is both too complex and massive for our purposes. Therefore, brushless gimbal motors were chosen for our device. They exhibit a low kV rating and are more capable of providing higher torques at low speeds, compared to standard brushless motors used in robotics, which run at high speeds and require significant gear reduction for usable output torque and speeds. The gimbal motors also provide rapid acceleration, an added bonus for the augmented human ability of extremely fast drumming. 

\subsubsection{Drivetrain}
A single stage timing belt drive was chosen for our prosthetic as it can withstand the repeated shock loading common during drumming. Additionally, a belt drives adds a compliant element between the output and the motors. It is important to have low to zero backlash to facilitate smooth drumming and precise control. Having a belt drive allows the motors (the heaviest components of the system) to be placed close to the elbow, thereby, reducing fatigue and serving as a mechanical fuse. Although belts can suffer wear and tear and may fail over time due to extended use, they can easily be replaced and have a low gear ratio. The final gear ratio of our device is less than 3:1.

\subsubsection{Structure}
The primary structure is a unibody frame machined from a single billet of aluminum. Diagonal supports are cut into the frame and this helps in removing as much material as possible without sacrificing structural rigidity. A structure cut out of a single piece of aluminum also has the advantage of having less weight than one which requires several parts and is held together by heavy steel fasteners. Furthermore, no extra hardware is required to hold the frame together. The frame has an i-beam profile that is closed on each end for stiffness. The control electronics are integrated into the main structure with cavities and mounting features in the middle of the frame. Cutouts along the top and bottom give access to connectors for sensors, power, and control bus. The motors are located in a custom sheet metal enclosure at the back of the main structure, tucked underneath the carbon fiber socket and slightly offset downwards.

\subsubsection{Electronics}
The main microprocessor boards used in the device are proprietary boards from {\it Meka Robotics} and are mounted directly to the main structure. Advanced high precision motor control is facilitated by using high resolution optical encoders mounted to the rear of the motors. The boards are equipped with analog inputs for taking input from various sensors (EMG, potentiometers) and also comes with an on-board 9-axis accelerometer. The communication between the host computer and the board is over a high speed {\it EtherCat} based control bus. The development platform ({\it MATLAB/Simulink}) is extremely flexible and can be used to rapidly employ different types of control schemes. 

\subsection{Sensing}

The primary method of sensing is electromyography. Two sets of muscles on the residual limb were chosen as the principal sensory inputs to the prosthesis. The action of playing a drumset involves several muscle groups throughout the entire body. We selected the extensor carpi ulnaris and flexor carpi radialis as these muscles do not play a significant role in the initial onset of a stroke (the up-down motion of the hand), thus, limiting cross-talk with other muscles used during drumming. 

These muscles are used to generate control signals that vary the parameters of the PID control system for the motors. Therefore, it becomes important that the system is fairly robust against false positives, otherwise, unwanted changes in the stick rebound will affect the quality of the musical output. 

\begin{figure}[ht]
\center
\includegraphics[width=.8\textwidth]{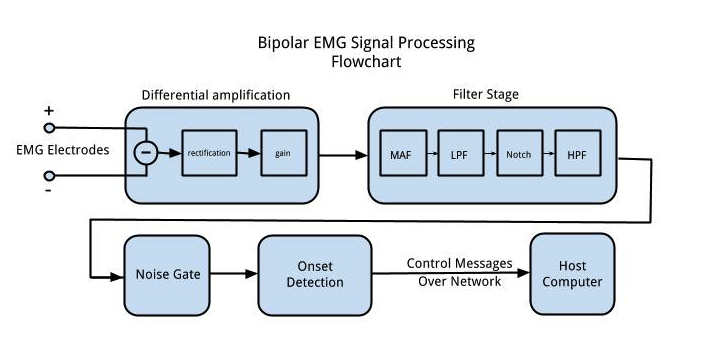}
\caption{EMG signal flow and processing.}
\label{EMGFlowDiagram}
\end{figure}

For each muscle group, a bipolar EMG recording paradigm was adopted. Surface electrodes were used and with the ground electrode attached to the bony region directly below the elbow. It is typical of commercial prostheses to use a hardware based differential amplifiers for bipolar EMG recordings. However, due to weight and general bulkiness of such amplifiers we perform differential amplification in software. The EMG electrodes are connected to a computer using a standard audio interface with built in pre-amplification. Following pre-amplification differential amplification, full wave rectification, and a gain multiplication are performed on the raw EMG input (see figure 3).

An RMS-based moving average filter (MAF) is applied on the amplified signal in order to smooth the signal by removing unwanted high frequency noise content. This stage is followed by three bi-quad filters whose coefficients are set in such a way that they act as a low pass filter (with cutoff of 520Hz), a notch filter (with center frequency around 180Hz and a Q-factor of approximately 1.5) and a high pass filter (with a cutoff frequency of approximately 20-30hz). The EMG muscle activations typically do not contain frequencies above 500Hz and the notch filter is used to reject the power-line interference. Although this may degrade signal quality, we are not as interested in power spectral density measures and the degradation has little effect on the onset detection methods. In addition to the standard power-line interference, electromagnetic noise due to the motors and on-board electronics also pose a big challenge for clean signal acquisition. The high pass filter is used to eliminate motion artifacts that arise due to the electrodes moving with respect to the skin or cable motion. A noise gate is used after this stage to remove low level residual noise as it further helps in improving the robustness of the onset detection system.

The real-time onset detection algorithm employed in this system is based on a bounded-Q filter-bank decomposition of the incoming signal and this has far superior performance than traditional envelope follower based onset detection methods \citep{puckette1998real}. This method is useful for percussion-like transients which do not lend themselves to sinusoidal decomposition. Various parameters such as the minimum velocity and the minimum temporal distance between the onsets can be adjusted as required. The detected onsets are mapped to various functionalities such as triggering single strikes of the stick, incrementally varying the parameters of the PID motor control system etc. Other features, such as the slowly evolving envelope contour of the EMG signal can easily be extracted, but due to the non-stationary nature of bio-signals in general, a meaningful mapping of continuously varying features to low-level parameters of the system is difficult. 

We evaluated the EMG filtering and onset detection system by constructing a labeled dataset based on Jason's muscle activations. Time series recordings from the two principal muscle locations were collected. When Jason contracted one of the muscles he provided a verbal cue indicating a perceived onset had occurred. These cues were used to label the recordings, which were then used for an objective evaluation. We achieved an F-measure of {\it F1 = .91}. Additionally, Jason provided subjective validation by designating the detection accuracy as sufficient for his needs.

For the purposes of the experiment that was designed to evaluate the entire prosthesis system, the onsets detected were used to change the proportional gain ({\it kp}) of the PID controller system in an incremental fashion. The useful range of the proportional gain parameter was determined empirically and the EMG onsets were used to change the parameter in a smooth linear fashion within this range.

\subsection{Evaluation}
The efficacy and success of a shared control system is typically evaluated by comparing the amount of effort required of the user with and without the system. For example, one may measure the time taken to complete a task \citep{carlson2013brain} or quantify the task difficulty and cognitive load involved using questionnaire forms \citep{gentili2013human}. We evaluate our system by measuring Jason's performance of a musical synchronization task. Our hypothesis is that by using a variable impedance system to control the stick tension, Jason will be able to achieve the different stroke types required of accomplished drummers. In this situation the initial onsets are achieved through the physical up and down motion of Jason's elbow. The bounce behavior, as a result of the stick tension, is controlled by the motor system and PID controller. Jason is able to manipulate the PID parameters using his muscles through EMG control.

\subsubsection{Experiment Procedures and Methods}
To test our hypothesis Jason was asked to play several common rhythmic motifs that require different types of strokes and bounce behaviors using the prosthetic. The objective is not to examine Jason's proficiency as a musician, but rather whether the electromechanical prosthetic design is sufficient for the needs of a drummer. In other words, does it effectively simulate the function of a drummer's fingers and grip in order to generate different stroke types? Additionally, is Jason capable of setting the proper PID parameters in order to generate the desired stroke type using the EMG interface? Therefore, the rhythmic patterns Jason was asked to play were patterns with which he was very familiar and had practiced for many hours both before and after his amputation (using a spring loaded drumming prosthetic he designed himself).

A 2x1 within subject study was conducted. Jason performed the study under two conditions: 1) using his own spring-loaded prosthetic and 2) using the new electromechanical prosthetic with variable impedance. The spring-loaded prosthetic functions similarly to our electromechanical prosthetic, however, instead of a motor simulating the grip to achieve a bounce, a spring is used. This prosthetic is the primary drumming tool Jason currently uses and has been using since his amputation. We chose to compare our prosthetic to the spring-loaded prosthetic as opposed to comparing the results to what other able-bodied drummers are able to achieve because a between subjects test would make it difficult to differentiate between the affects of the prosthetic and differences in participants' musical proficiencies. Likewise, comparing results between Jason's biological left arm and our new prosthetic may also simply distinguish the natural strengths and weaknesses between his two arms. Drummers often have a dominant hand, which can outperform the other.

Therefore, we chose to compare the electromechanical prosthetic directly to Jason's spring-loaded prosthetic. We are confident that in doing this we can directly evaluate the effect our prosthetic has on Jason's performance abilities. Additionally, Jason is already considered a proficient drummer when using his spring-loaded prosthetic setting the musical baseline of our electromechanical design at a performance level similar to professional musicians. Despite Jason's skill with his spring-loaded prosthetic, there are limitations of using a spring to generate bounce, hence, our motivation for a new design. Any improvements as a result of our prosthetic might be small, but are essential for becoming a more well-rounded and accomplished musician.

The prime issue with Jason's prosthetic is that the spring's tension cannot be manipulated, thus, only a single bounce behavior is possible. Variable bounce behaviors will allow for double strokes with varying intervals making our prosthetic more amenable to a wider range of tempi. To test this hypothesis Jason completed rhythmic synchronization tasks based on the common rhythmic drumming patterns. The notion is that Jason will be able to more accurately and consistently play rhythmic motifs if the rebound of the stick behaves in the appropriate manner. This will result in better synchronization between Jason's play and the perfectly timed reference audio. The rhythmic motifs were generated using a single audio drum sample that was arranged according to the appropriate temporal patterns using the computer's internal clock. Five patterns were chosen and each was played at tempi starting from 90bpm to 210bpm increasing at 10bpm intervals. The procedure was as follows:

\begin{enumerate}
  \item Listen to computer audio of rhythmic motif for two measures
  \item Perform four measures of the motif while trying to sync perfectly with the audio file (and record audio of performance)
  \item Repeat steps 1 and 2 for a wide range of tempos
  \item Repeat steps 1-3 for each of the five motifs
\end{enumerate}

This procedure was performed under two conditions: 1) using the electromechanical prosthetic and 2) using the spring-loaded prosthetic.

\subsubsection{Results}
We analyzed the audio recordings to each of the rhythmic motif performances for the varying tempi. In order to evaluate synchronicity we used a dynamic time warping (DTW) algorithm. DTW is an algorithm that measures similarity between two time series. We compared the audio of Jason's playing with that of the computer generated audio track. The resulting cost of alignment calculated by DTW provides a distance-like metric. For two identical time series this distance is zero, but as the sequences vary the distance increases. Therefore, DTW can be used as a tool for measuring synchronization \citep{suneffect}.

\begin{figure}[ht!]
\center
\includegraphics[width=.8\textwidth]{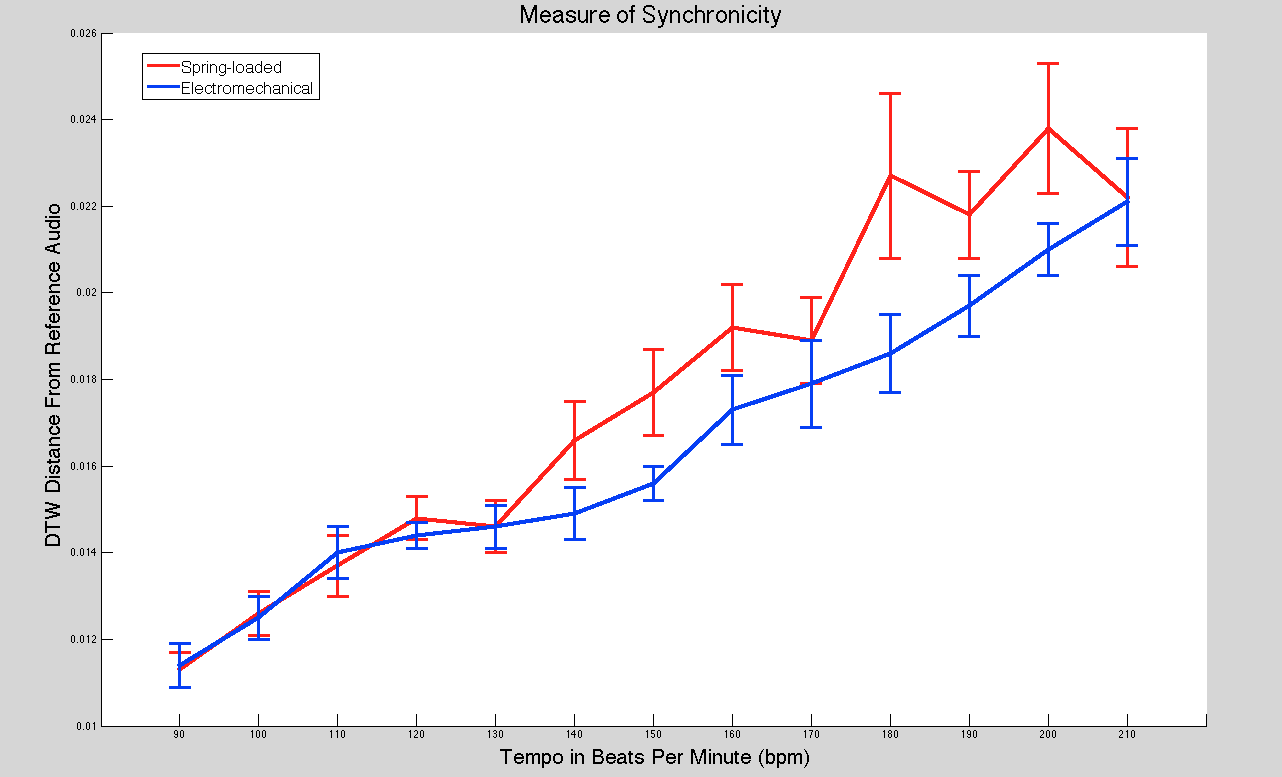}
\caption{A comparison of the ability to synchronize with a reference audio using the spring-loaded and electromechanical prostheses over a range of tempi. Dynamic time warping is used to measure the distance between the reference and recorded performance. A smaller DTW distance suggests the performance and reference are more closely aligned. }
\label{synchronicity}
\end{figure}

\begin{table}[h!]
\centering
  \caption{Average distances between the live performances using the spring-loaded and electromechanical prostheses and reference audio for each tempo based on a DTW metric. A {\it t}-test was used to test for significant differences and is indicated at the p\textless.05 level. A Bonferroni correction was used to control for multiple comparisons.} 
\begin{tabular}{| c | c | c | c |}
    \hline
    BPM & Spring-loaded & Electromechanical & 95\% Confidence \\ \hline
    \hline
    90 & .0113 & .0114 &  \\ \hline
    100 & .0126 & .0125 &  \\ \hline
    110 & .0137 & .0140 &  \\ \hline
    120 & .0148 & .0144 &  \\ \hline
    130 & .0146 & .0146 &  \\ \hline
    140 & .0166 & .0149 & * \\ \hline
    150 & .0177 & .0156 & * \\ \hline
    160 & .0192 & .0173 &  \\ \hline
    170 & .0189 & .0179 & * \\ \hline
    180 & .0227 & .0186 & * \\ \hline
    190 & .0218 & .0197 & * \\ \hline
    200 & .0238 & .0210 & * \\ \hline
    210 & .0222 & .0221 &   \\\hline
     \multicolumn{4}{c}{}  
\end{tabular}
\end{table}

Synchronicity was measured for each of the rhythmic motifs for all the tempos. The normalized costs were averaged across the five motifs for each of the 12 tempi. The figure shows these averaged results from 90bpm to 210bpm. {\it T}-tests were used to test for significant differences for each tempo and a Bonferroni correction of the {\it p}-values was performed to control for multiple comparisons. Table 1 shows the complete results.

\subsubsection{Discussion}
As predicted the synchronization improved over a range of tempi using the electromechanical prosthetic. The absence of significant differences from 90bpm - 140bpm is to be expected because these are speeds at which doubles strokes are unnecessary allowing Jason to use single strokes. In other words, the speed at which Jason can move his elbow up and down is sufficient to achieve decent synchronization.

At rates faster than 140bpm double strokes become necessary to accurately play the pattern and this is where the performance of the electromechanical and spring-loaded prostheses begin to diverge. The decrease in synchronization remains roughly linear for the electromechanical prosthetic over the course of all the tempi. The increased performance at 210bpm for the spring-loaded prosthetic is most likely due to the spring being naturally sufficient for double bounce intervals at that tempo. Though, as is evident, variable spring tension is necessary for sufficient double bounces of other intervals.

The results are promising and we believe more improvement can be expected. Jason had roughly 20 hours of playing time with the electromechanical prosthetic over the course of four months prior to this study. This is compared to hundreds of hours of playing time over the course of two years with the spring loaded prosthetic. Additionally, this study required Jason to play only four measures, thus, not allowing for a decent evaluation of endurance. We predict that a better ability to perform these motifs for longer periods of time would also become apparent if measured, as double strokes allow drummers to conserve energy and prevent muscle fatigue.

\section{A Wearable Robotic Musician: Incorporating a Second Stick}
In addition to providing increased musical virtuosity for an amputee, the advent of a wearable electromechanical device also permits the exploration for other musical endeavors. One such endeavor is the notion of a wearable robotic musician. In this section we discuss the addition of a second stick to the prosthesis that behaves autonomously as if it were an individual agent. The second stick follows the same hardware blueprint of the primary stick, however, it is programmed to behave in a much different manner. We discuss the interesting implications on both human-robotic interaction and musical performance that such a design engenders by describing several behaviors developed for the agent.

\begin{figure}[ht]
\center
\includegraphics[width=.8\textwidth]{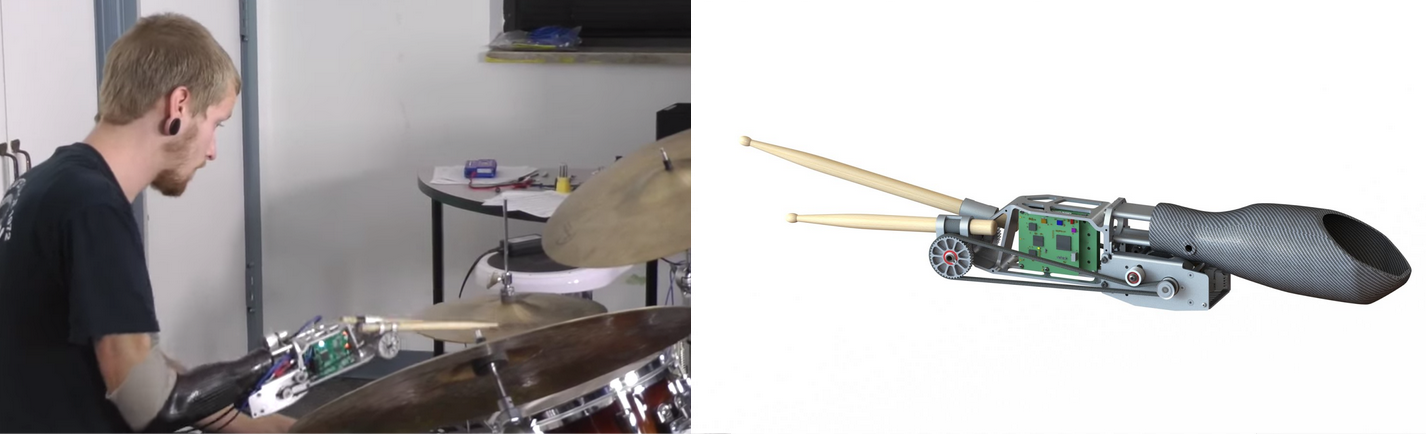}
\caption{A second stick is included that has a ``mind of its own" and creates rhythmic responses to Jason's and other musicians playing.}
\label{secondary}
\end{figure}

\subsection{A New Paradigm of Robotic Musicianship}
Robotics in music have primarily come in two forms: 1) independent and stand-alone agents \citep{hoffman2011interactive} and 2) augmented instruments in which mechanics are mounted to traditional instruments resulting in increased or modified functionality \citep{kapur2005history, richardGreene}. Human musicians interact with systems of the first form similarly to how human musicians interact with each other within an ensemble. There's a significant degree of independence, however, there is a shared and agreed upon musical context (such as tempo, time signature, key signature, and genre) that each musician has a responsibility to adhere to or at least address. Each musician may also have a specific role (such as keeping the beat, providing harmonic foundation, or improvising) that allows them to contribute to the higher-level musical product of the group.

The interaction between a person and an augmented robotic musical instrument may support standard playing technique or may require the person to approach the instrument in a non-traditional manner. Both paradigms present an interesting interaction model resulting from a shared acoustic interface. As the person performs on the instrument, the robotic components generate a response based on something it senses about the music, person, or environment. The robot may be programmed to behave similarly to that of stand-alone musical agents, however, the shared interface allows the person to influence the robot's final sonic result by physically manipulating certain aspects of the instrument (such as muting resonators). Likewise, the robot may similarly influence the person's sound and playing through analogous physical means. 

A wearable robotic musician extends the notion of a shared interface to that of a shared physical actuator and manipulator. Not only is the instrument shared, but also one or more of the DoFs necessary for actuating the instrument is shared. In the case of this prosthetic, the second stick can be used in two ways. It may behave similarly to the primary stick in that it bounces in a particular manner based on autonomously set PID settings. In this situation the user must trigger the initial onset. Subsequent onsets, resulting from the stick's bounce behavior, are controlled by the artificial intelligence (AI). The AI does not attempt to produce the desired effect expressed by the user, but rather an effect it computes to be appropriate based on what it senses about the music and the person's playing.

\begin{figure}[ht]
\center
\includegraphics[width=.8\textwidth]{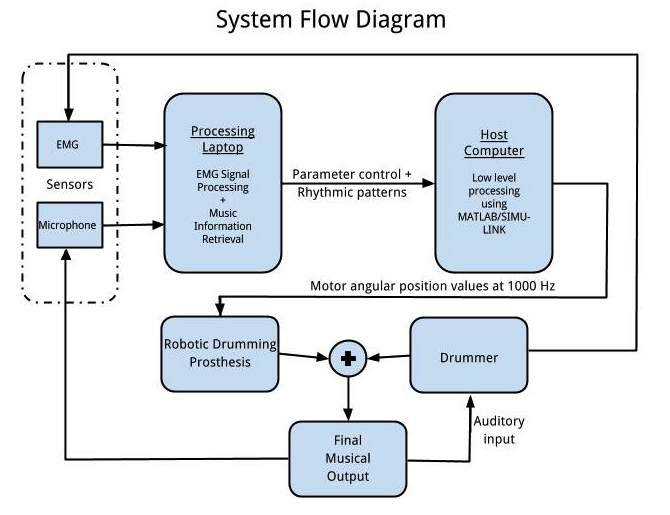}
\caption{}
\label{systemflow}
\end{figure}

A second manner in which the robotic musician may behave is by generating its own strikes. In this scenario, the motor is used to both drive initial onsets (functioning like an elbow) and control the bounce behavior. However, the user's elbow DoF controls the distance between the sticks and the drum-head surface. This means the user has control over volume and whether contact is made with the drum at all. From an aesthetic and musical standpoint this scenario is much more interesting because the AI can generate rhythms and timbres that play a larger role on the final musical output than that of merely bounce behavior. This interaction paradigm is an example of haptic shared control in which the user and robot directly provide input to the music (see figure 6).

\subsection{Behaviors of the Wearable Robotic Musician}
Several behaviors were designed for the autonomous secondary stick that enabled different interactive schemes between human and robot such as turn-taking and accompaniment. These behaviors were designed based on aesthetics and what we thought had the potential to create interesting musical results. These behaviors were implemented and showcased at various locations around the world. We invite readers to view a concert conducted at the Bailey Performance Center at Kennesaw State University as part of the Atlanta Science Festival featuring many of these behaviors at \url{https://www.youtube.com/watch?v=dlSZCu5FAVM}.

One behavior uses a note generative model based on rhythms provided by other musicians. The system analyzes the audio offline and extracts patterns based on rhythm and dynamics. Specifically, we used patterns extracted from excerpts of a percussion performance of renowned physicist, Richard Feynman. The secondary stick then generates these rhythmic sequences and the user can interact with them by layering the rhythms with accompanying sounds using his other limbs as well as providing alternative accent placements by manipulating distance between the electromechanical arm and drum surface.

A second behavior is based on inter-musician communication. The autonomous stick listens to the playing of other musicians in the group and generates responses according to their play. Specifically, the stick listens to the chords played by a guitarist and adjusts its speed based on the chords' tonal centers. The stick is capable of generating strikes at a rate of up to 20Hz. When using such high speeds the acoustic response perceived by the ear is more timbral in nature rather than rhythmic (akin to different sounding buzz rolls). As the guitarist changes chords the strike frequency changes within the range such that the acoustic timbre is manipulated.

A third behavior uses EMG and allows the user to control higher-level parameters regarding the sticks musical response. One higher-level parameter we explored is note density (the number of onsets within a time frame). In this behavior, the user provides a ``seed" rhythm by flexing his muscle rhythmically. The system uses the onset detection method described earlier to learn the temporal sequence and quantizes it based on the desired tempo. The robot then stochastically adds additional notes to the rhythm to increase note density based on the user's control. In order to control the note density the user flexes continuously in which a stronger amplitude is mapped to a higher note density.

Robotic musicianship behaviors such as these are difficult to evaluate objectively. Often, a more subjective overview of the challenges and procedures used to create interesting behaviors is more helpful for those developing generative algorithms. As such, we have empirically comprised a list of considerations we believe can be useful for other designers developing such a wearable robotic musician paradigm based on Jason's and other musician's experience of performing and interacting with the system.

\begin{description}
  \item [Do not replace the human drummer] It is important that the robotic musician does cause the human drummer to relinquish the role of his arm entirely to the AI. Rather, design algorithms that supplement or complement the drummer's play.
  
  \item [Consider the natural abilities of the drummer] Generative algorithms based on outputs that are outside the reach of the drummer are preferred. The stick's ability of moving at speeds greatly exceeding that of natural human ability and performing complex rhythms accurately should be leveraged. For example, allowing the robot to perform the complex Feynman rhythms enabled Jason to use his other limbs to create a richer musical outcome.
  
  \item [Use the robot sparingly] It is important to find the appropriate balance such that the drummer does not find the robotic musician to be an interference, but rather an inspiration to his and other musician's in the group.
  
  \item [Consider cognitive load on the drummer] If the drummer is controlling aspects of the robot's behaviors the mappings should be based on higher-level musical parameters (such as note density).
  
\end{description}

These are a few considerations we found necessary to address in working with Jason and the prosthetic. However, we have only begun to explore the possibilities of wearable robotic musicians and believe there is great potential for further research and development in this area.

\section{Conclusion}
In this work we presented a drumming prosthesis for a musician with a transradial amputation. The mechanical design and underlying computational architecture created an effective simulation of finger and grip function during drumming applications. The system was validated empirically by discussions with the user and objectively through a user study involving a musical synchronization task.

Developing prostheses that enable people to play musical instruments is very challenging because the necessary level of dexterity, control, and feel afforded to the user is enormous. A perfect replication of natural wrist and hand anatomy is ideal, however, alternative methods can be leveraged to circumvent the obvious difficulties of achieving this. A shared control framework is helpful because it allows the user to achieve responses and behaviors in the device without having to cognitively address each individual parameter or DoF. Instead, the user is able to provide a higher level input, which the robot interprets and uses to manipulate parameters in order to generate the desired response.

The prospect of a wearable robotic musician was also discussed in this article. We explored different interactive scenarios in which musicians perform with the robot enabling shared control over the final artistic output. Our empirical findings are not only useful for composers and musicians, but also for researchers developing wearable artificial supernumerary limbs. Understanding how one interacts with a wearable autonomous agent is important for optimizing coordination tasks and user experience. Supernumerary limbs have the potential to augment human anatomy in several ways. This may include the ability to surmount physical limitations or, as in our case, to encourage creativity and decision processes by presenting the user with new ideas, interpretations, or solutions.

\section{Acknowledgements}
This work was supported by NSF Grant No. 1345006. The authors thank Roberto Aimi, Minwei Gu, Guy Hoffman, Rob Kistenberg, Iman Mukherjee, and Annie Zhang for their valuable insight, suggestions, and contributions to this work.

%\section{Biblography}
\bibliography{main}

\begin{thebibliography}{}

\bibitem[\protect\citeauthoryear{Abbink and
  Mulder}{2010}]{abbink2010neuromuscular}
David~A Abbink and M~Mulder.
\newblock Neuromuscular analysis as a guideline in designing shared control.
\newblock {\em Advances in haptics}, 109:499--516, 2010.

\bibitem[\protect\citeauthoryear{Cakmak \bgroup \em et al.\egroup
  }{2010}]{cakmak2010designing}
Maya Cakmak, Crystal Chao, and Andrea~L Thomaz.
\newblock Designing interactions for robot active learners.
\newblock {\em Autonomous Mental Development, IEEE Transactions on},
  2(2):108--118, 2010.

\bibitem[\protect\citeauthoryear{Carlson and
  Mill{\'a}n}{2013}]{carlson2013brain}
Tom Carlson and Jos{\'e} del~R Mill{\'a}n.
\newblock Brain-controlled wheelchairs: a robotic architecture.
\newblock {\em IEEE Robotics and Automation Magazine},
  20(EPFL-ARTICLE-181698):65--73, 2013.

\bibitem[\protect\citeauthoryear{Cipriani \bgroup \em et al.\egroup
  }{2008}]{cipriani2008shared}
Christian Cipriani, Franco Zaccone, Silvestro Micera, and Maria~Chiara
  Carrozza.
\newblock On the shared control of an emg-controlled prosthetic hand: analysis
  of user--prosthesis interaction.
\newblock {\em Robotics, IEEE Transactions on}, 24(1):170--184, 2008.

\bibitem[\protect\citeauthoryear{Cipriani \bgroup \em et al.\egroup
  }{2009}]{cipriani2009progress}
Christian Cipriani, Marco Controzzi, and M~Chiara Carrozza.
\newblock Progress towards the development of the smarthand transradial
  prosthesis.
\newblock In {\em Rehabilitation Robotics, 2009. ICORR 2009. IEEE International
  Conference on}, pages 682--687. IEEE, 2009.

\bibitem[\protect\citeauthoryear{Davenport}{2013}]{davenport2013supernumerary}
Clark Clark~Michael Davenport.
\newblock {\em Supernumerary robotic limbs: biomechanical analysis and
  human-robot coordination Training}.
\newblock PhD thesis, Massachusetts Institute of Technology, 2013.

\bibitem[\protect\citeauthoryear{Dellon and
  Matsuoka}{2007}]{dellon2007prosthetics}
Brian Dellon and Yoky Matsuoka.
\newblock Prosthetics, exoskeletons, and rehabilitation [grand challenges of
  robotics].
\newblock {\em Robotics \& Automation Magazine, IEEE}, 14(1):30--34, 2007.

\bibitem[\protect\citeauthoryear{Gentili \bgroup \em et al.\egroup
  }{2013}]{gentili2013human}
Rodolphe~J Gentili, Hyuk Oh, Isabelle~M Shuggi, Ronald~N Goodman, Jeremy~C
  Rietschel, Bradley~D Hatfield, and James~A Reggia.
\newblock Human-robotic collaborative intelligent control for reaching
  performance.
\newblock In {\em Foundations of Augmented Cognition}, pages 666--675.
  Springer, 2013.

\bibitem[\protect\citeauthoryear{Herr \bgroup \em et al.\egroup
  }{2003}]{herr2003cyborg}
Hugh Herr, Graham~Paul Whiteley, and Dudley Childress.
\newblock {\em Cyborg Technology--Biomimetic Orthotic and Prosthetic
  Technology}.
\newblock SPIE Press, Bellingham, Washington, 2003.

\bibitem[\protect\citeauthoryear{Hochberg \bgroup \em et al.\egroup
  }{2006}]{hochberg2006neuronal}
Leigh~R Hochberg, Mijail~D Serruya, Gerhard~M Friehs, Jon~A Mukand, Maryam
  Saleh, Abraham~H Caplan, Almut Branner, David Chen, Richard~D Penn, and
  John~P Donoghue.
\newblock Neuronal ensemble control of prosthetic devices by a human with
  tetraplegia.
\newblock {\em Nature}, 442(7099):164--171, 2006.

\bibitem[\protect\citeauthoryear{Hoffman and
  Weinberg}{2010}]{hoffman2010shimon}
Guy Hoffman and Gil Weinberg.
\newblock Shimon: an interactive improvisational robotic marimba player.
\newblock In {\em CHI'10 Extended Abstracts on Human Factors in Computing
  Systems}, pages 3097--3102. ACM, 2010.

\bibitem[\protect\citeauthoryear{Hoffman and
  Weinberg}{2011}]{hoffman2011interactive}
Guy Hoffman and Gil Weinberg.
\newblock Interactive improvisation with a robotic marimba player.
\newblock {\em Autonomous Robots}, 31(2-3):133--153, 2011.

\bibitem[\protect\citeauthoryear{Kapur \bgroup \em et al.\egroup
  }{2011}]{kapur2011machine}
Ajay Kapur, Michael Darling, Dimitri Diakopoulos, Jim~W Murphy, Jordan
  Hochenbaum, Owen Vallis, and Curtis Bahn.
\newblock The machine orchestra: An ensemble of human laptop performers and
  robotic musical instruments.
\newblock {\em Computer Music Journal}, 35(4):49--63, 2011.

\bibitem[\protect\citeauthoryear{Kapur}{2005}]{kapur2005history}
Ajay Kapur.
\newblock A history of robotic musical instruments.
\newblock In {\em Proceedings of the International Computer Music Conference},
  pages 21--28. Citeseer, 2005.

\bibitem[\protect\citeauthoryear{Kim \bgroup \em et al.\egroup
  }{2006}]{kim2006continuous}
Hyun~K Kim, J~Biggs, David~W Schloerb, Jose~M Carmena, Mikhail~A Lebedev,
  Miguel~AL Nicolelis, and Mandayam~A Srinivasan.
\newblock Continuous shared control for stabilizing reaching and grasping with
  brain-machine interfaces.
\newblock {\em Biomedical Engineering, IEEE Transactions on}, 53(6):1164--1173,
  2006.

\bibitem[\protect\citeauthoryear{Lake and Dodson}{2006}]{lake2006progressive}
Chris Lake and Robert Dodson.
\newblock Progressive upper limb prosthetics.
\newblock {\em Physical medicine and rehabilitation clinics of North America},
  17(1):49--72, 2006.

\bibitem[\protect\citeauthoryear{Li \bgroup \em et al.\egroup
  }{2011}]{li2011dynamic}
Qinan Li, Weidong Chen, and Jingchuan Wang.
\newblock Dynamic shared control for human-wheelchair cooperation.
\newblock In {\em Robotics and Automation (ICRA), 2011 IEEE International
  Conference on}, pages 4278--4283. IEEE, 2011.

\bibitem[\protect\citeauthoryear{Llorens-Bonilla \bgroup \em et al.\egroup
  }{2012}]{llorens2012demonstration}
Baldin Llorens-Bonilla, Federico Parietti, and H~Harry Asada.
\newblock Demonstration-based control of supernumerary robotic limbs.
\newblock In {\em Intelligent Robots and Systems (IROS), 2012 IEEE/RSJ
  International Conference on}, pages 3936--3942. IEEE, 2012.

\bibitem[\protect\citeauthoryear{Logan-Greene}{}]{richardGreene}
Richard Logan-Greene.
\newblock The music of richard johnson logan-greene.
\newblock \url{http://zownts.com}.
\newblock Accessed: 5-1-2014.

\bibitem[\protect\citeauthoryear{Maes \bgroup \em et al.\egroup
  }{2011}]{maes2011man}
Laura Maes, Godfried-Willem Raes, and Troy Rogers.
\newblock The man and machine robot orchestra at logos.
\newblock {\em Computer Music Journal}, 35(4):28--48, 2011.

\bibitem[\protect\citeauthoryear{Nudehi \bgroup \em et al.\egroup
  }{2005}]{nudehi2005shared}
Shahin~S Nudehi, Ranjan Mukherjee, and Moji Ghodoussi.
\newblock A shared-control approach to haptic interface design for minimally
  invasive telesurgical training.
\newblock {\em Control Systems Technology, IEEE Transactions on},
  13(4):588--592, 2005.

\bibitem[\protect\citeauthoryear{Puckette \bgroup \em et al.\egroup
  }{1998}]{puckette1998real}
Miller~S Puckette, Theodore Apel, et~al.
\newblock Real-time audio analysis tools for pd and msp.
\newblock 1998.

\bibitem[\protect\citeauthoryear{Schirner \bgroup \em et al.\egroup
  }{2013}]{schirner2013future}
Gunar Schirner, Deniz Erdogmus, Kaushik Chowdhury, and Taskin Padir.
\newblock The future of human-in-the-loop cyber-physical systems.
\newblock 2013.

\bibitem[\protect\citeauthoryear{Singer \bgroup \em et al.\egroup
  }{2004}]{singer2004lemur}
Eric Singer, Jeff Feddersen, Chad Redmon, and Bil Bowen.
\newblock Lemur's musical robots.
\newblock In {\em Proceedings of the 2004 conference on New interfaces for
  musical expression}, pages 181--184. National University of Singapore, 2004.

\bibitem[\protect\citeauthoryear{Sun \bgroup \em et al.\egroup
  }{2012}]{suneffect}
Sisi Sun, Trishul Mallikarjuna, and Gil Weinberg.
\newblock Effect of visual cues in synchronization of rhythmic patterns.
\newblock 2012.

\bibitem[\protect\citeauthoryear{Weinberg and
  Driscoll}{2006a}]{weinberg2006toward}
G.~Weinberg and S.~Driscoll.
\newblock Toward robotic musicianship.
\newblock {\em Computer Music Journal}, 30(4):28--45, 2006.

\bibitem[\protect\citeauthoryear{Weinberg and
  Driscoll}{2006b}]{weinberg2006robot}
Gil Weinberg and Scott Driscoll.
\newblock Robot-human interaction with an anthropomorphic percussionist.
\newblock In {\em Proceedings of the SIGCHI conference on Human Factors in
  computing systems}, pages 1229--1232. ACM, 2006.

\bibitem[\protect\citeauthoryear{Wu and Asada}{2014}]{wu2014bio}
Faye~Y Wu and Harry Asada.
\newblock Bio-artificial synergies for grasp posture control of supernumerary
  robotic fingers.
\newblock 2014.

\end{thebibliography}
\bibliographystyle{named}

%\hrule
%\vspace*{.1in}
%Authors' contact information: Mason Bretan, Center for Music Technology, Georgia Tech, Atlanta, GA.  Email: pbretan3@gatech.edu.  Deepak Gopinath, Center for Music Technology, Georgia Tech, Atlanta, GA.  Email: deg3@gatech.edu. Philip Mullins, Meka Robotics, San Francisco, CA.  Email: philip.mullins@google.com. Gil Weinberg, Center for Music Technology, Georgia Tech, Atlanta, GA.  Email: gilw@gatech.edu.

\end{document}